\documentclass[aps,preprint,amsmath,amssymb,amsfonts,nofootinbib,superscriptaddress]{revtex4-1}
\usepackage{epsfig}
\usepackage{graphicx}
\usepackage{dcolumn}
\usepackage{bm}
\usepackage{amsthm}
\usepackage{amsmath}
\usepackage{color}
\usepackage{natbib} 
\usepackage{graphicx}
\usepackage[export]{adjustbox}
\usepackage{amssymb}
\usepackage{amsmath}
\usepackage{amsfonts}
\usepackage{amssymb}
\usepackage{braket}
\usepackage{xcolor}
\usepackage{subcaption}

\usepackage[hidelinks]{hyperref}
\hypersetup{
    colorlinks=false,
    linkcolor=false,
    filecolor=false,      
    urlcolor=false,
    pdfpagemode=FullScreen,
    }
\usepackage{float}

\newcommand{\beq}{\begin{equation}}
\newcommand{\eeq}{\end{equation}}
\newcommand{\bea}{\begin{eqnarray}}
\newcommand{\eea}{\end{eqnarray}}

\newcommand{\eq}{\begin{equation}}
\newcommand{\eqx}{\end{equation}}
\newcommand{\eqn}{\begin{eqnarray}}
\newcommand{\eqnx}{\end{eqnarray}}

\newcommand{\f}[2]{\frac{#1}{#2}}

\begin{document}

\title{Neural Network Complexity of Chaos and Turbulence}
\author{Tim Whittaker}%
\affiliation{
Département des sciences de la Terre et de l'atmosphère, Université du Québec à Montréal, Montréal, QC H3C 3P8, Canada \\ whittaker.tim@courrier.uqam.ca
}\email{Corresponding Author}

\author{Romuald A. Janik}
\affiliation{
Institute of Theoretical Physics and Mark Kac Center for Complex Systems Research,
Jagiellonian University,
ul. {\L}ojasiewicza 11, 30-348 Krak{\'o}w, Poland \\ romuald.janik@gmail.com
}

\author{Yaron Oz}%
\affiliation{
 Raymond and Beverly Sackler School of Physics and Astronomy, Tel-Aviv University, Tel-Aviv 69978, Israel. \\ yaronoz@tauex.tau.ac.il
}
\affiliation{Simons Center for Geometry and Physics, SUNY, Stony Brook, NY 11794, USA}
%


%
\date{\today}

\begin{abstract}
Chaos and turbulence are complex physical phenomena, yet a precise definition
of the complexity measure that quantifies them is still lacking. In this
work we consider the relative complexity of chaos and turbulence from
the perspective of deep neural networks.
We analyze a set of classification problems, where the network has to distinguish images of fluid profiles in the turbulent regime from other classes of images such as fluid profiles in the chaotic regime, various constructions of noise and real world images. 
We analyze incompressible as well as weakly compressible fluid flows.
We quantify the complexity of the computation performed by the network
via the intrinsic dimensionality of the internal feature representations, and calculate the effective number of independent features which the network uses
in order to distinguish between classes.
In addition to providing a numerical estimate of the complexity of the computation, the measure also characterizes the neural network processing at intermediate and final stages.
We construct adversarial examples and use them to identify
the two point correlation spectra for the chaotic and turbulent vorticity
as the feature used by the network for classification.
\end{abstract}

\maketitle
\tableofcontents

\section{Introduction}

Fluid turbulence is a major unsolved problem of physics exhibiting an 
emergent complex structure from simple rules, that is Newton's second law applied to the fluid elements
(for a review see e.g. \cite{Frisch}).
Most fluid motions
in nature at all scales are turbulent, yet despite centuries of research, we still lack
a complete understanding of fluid flows in the non-linear regime.
Insights into turbulence are likely to offer guidance to the principles and
dynamics of  non-linear systems with a large number of strongly interacting degrees of freedom far from equilibrium.

An important dimensionless parameter in the study of fluid flows  is the Reynolds number 
${\cal R}_e = \frac{l v}{\nu}$, where $l$
is a characteristic length scale, $v$ is the velocity difference at that scale, and $\nu$ is the kinematic viscosity. 
The Reynolds number quantifies the relative strength of the
non-linear interaction compared to the viscous term.
When the Reynolds number is of order $10-10^2$ one observes a chaotic 
fluid flow, while when it is $10^3$ or higher,
one observes a fully developed turbulent structure of the flow.
The turbulent velocity field exhibits highly complex spatial and temporal structures and  appears to be a random process.

A single realization of a turbulent solution to the NS equations is unpredictable even in the absence of a random force. However, studying statistical averages seems to reveal a hidden
scaling structure.
Indeed, experimental and numerical  data suggest that turbulent fluid flows exhibit a statistically homogeneous and isotropic steady state at the inertial range of scales $l \ll r \ll L_f$, where the distance scales $l$ and $L_f$ are determined in terms of the viscosity and driving force, respectively. The properties of the this statistical structure can be quantified by 
studying averages of fluid observables.
For instance, 
if we denote the velocity of the fluid by $\vec{v}(t,\vec{r})$ then the turbulent behavior can be characterized by the longitudinal structure functions $S_n(r) = \langle (\delta v(r))^n \rangle$ of velocity differences 
$\delta v(r)  = (\vec{v}(\vec{r}) - \vec{v}(0))\cdot \frac{\vec{r}}{|\vec{r}|}$
between points separated by a fixed distance $r$. In the inertial range of scales these correlation functions exhibit a universal scaling law:
\begin{equation}
    S_n(r) \sim r^{\xi_n} \ ,
\label{expo}
\end{equation}
where the exponents $\xi_n$  are independent of the fluid
details and depend only on the number of spatial dimensions.
On the other hand, the overall coefficients in (\ref{expo}) are not universal.

In a seminal work \cite{Kolmogorov}, Kolmogorov used the inertial range 
cascade-like behavior (introduced by Richardson)
of incompressible non-relativistic fluids, where large eddies break into smaller eddies in a process where energy is transferred without dissipation. 
Assuming scale invariant statistics for this direct cascade (from large to small length scales), he deduced that $\xi_n = n/3$. 
Thus, for instance, the Fourier transform of $S_2$ gives the energy power spectrum 
that exhibits Kolmogorov scaling:
\begin{equation}
    E(k) \sim k^{-\frac{5}{3}} \ .
    \label{EK}
\end{equation}
It is well known that Kolmogorov linear scaling is corrected by intermittency in all the direct cascades\footnote{$\xi_3=1$ is an exact result which can be derived analytically together with only a few other analytical cases \cite{Falkovich:2009mb}.}, and seems 
to be exact in the two-dimensional inverse cascade. In the inverse cascade the energy flows
from the UV to the IR and the inertial range is $r \gg L_f$.

The currently available experimental \cite{Benzi1995OnTS} and numerical \cite{chen_dhruva_kurien_sreenivasan_taylor_2005,Biferale_2019} methods provide a valuable, yet modest information about the 
anomalous scaling exponents $\xi_n$. The data shows a clear deviation of the scaling exponents
from Kolmogorov scaling in three space dimensions, but it is not sufficiently accurate 
\cite{Biferale_2019}
to distinguish between the various models 
\cite{PhysRevLett.72.336,PhysRevE.63.026307,Eling2015TheAS,Oz:2017ihc}, 
that have been proposed to explain them.
One would need to know higher $n$ exponents with a better precision in order
to gain further insight to the turbulence statistics and to the underlying dynamics
that explains them. In particular, one would like to have an access
to rare solutions that are important to probe the tail of the statistical distribution
and calculate the higher $n$ exponents.
Thus, we should aim for an era of ``precision turbulence".

Deep learning has revolutionized diverse fields by proving capable
of effective learning of statistical distributions, both in identifying data correctly as
well as generating new ones. 
The ability of deep networks to generalize
from learning data to test data as well as to generating new data samples
lead to an intensive research in recent years 
with the aim of characterizing, both theoretically and empirically,  the complexity measures that quantify the generalization capability of the networks. 
The neural network complexity measures are constructed such that lower complexity implies a smaller generalization gap (see \cite{https://doi.org/10.48550/arxiv.1912.02178} for the analysis of forty such complexity measures).
However, the notions of complexity addressing the generalization gap are not really adequate for our purposes, as we will be interested in quantifying the complexity of neural network computation as a function of depth (intuitively corresponding to scale).

While turbulence is evidently a complex phenomenon, a precise definition
of a complexity measure that quantifies it is still lacking. Among the attempts that have been
made we point out the use of Kolmogorov complexity to characterize its randomness
as well as a recent use of information theory in order to compare thermal equilibrium and turbulence \cite{Shavit_2020}.
In this respect, we would like to emphasize the differences between the current work and other applications of deep neural networks to turbulence. These include feature extraction \cite{doi:10.1080/14685248.2020.1757685, moghaddam2018deep}, turbulent flow classification \cite{li_yang_zhang_he_deng_shen_2020, Buzzicotti2022, PhysRevFluids3104604}, explainable feature extraction \cite{lellep_prexl_eckhardt_linkmann_2022}, generative models \cite{Drygala_2022,tretiak2022physicsconstrained,yang2023denoising,Shu_2023,mohan2019compressed,king2018deep}, super-resolution techniques \cite{fukami_fukagata_taira_2019,kim_kim_won_lee_2021, ClarkDiLeoni2023}, inpainting methods \cite{PhysRevFluids6050503} and 
learning the Navier-Stokes equations in order to generate fluid flows from new unseen initial conditions~\cite{li2020fourier, https://doi.org/10.48550/arxiv.2207.14366}
(see \cite{https://doi.org/10.1002/gamm.202100002, doi:10.1146/annurev-fluid-010518-040547, fluids7030116,doi:10.1146/annurev-fluid-010719-060214, Panchigar2022} for some comprehensive reviews). 
Our study has a different focus and aims compared to these works. Our goal is to quantify the computational complexity of distinguishing turbulence from chaotic dynamics at different length scales corresponding to different depths in the neural network and compare it with other classification tasks such as that of natural images. In order to do that we employ the complexity measure proposed in~\cite{complexity}, and apply it to a neural network classifier architecture applied to turbulence. While both turbulent flows and neural network architecture are quite standard, the neural network complexity measure provides a new insight into the underlying structure of turbulence.

We will analyze a set of classification problems where the network has to distinguish images of fluid profiles in the turbulent regime from other classes of images such as fluid profiles in the chaotic regime, various constructions of noise and real world images of cats or dogs.
We would like to gain insight into the relative difficulty of the above classification tasks in order to understand how different the turbulent images are from these other classes as viewed by the deep neural network.
Also, we would like to identify what features are being identified by the neural network to distinguish turbulent from chaotic fluid flows.
However, simply looking at the classification accuracy is insufficient, since as we will see in basically all of these cases, a standard neural network (ResNet-18 \cite{DBLP:journals/corr/HeZRS15}) can be trained to reach 99\% or 100\% accuracy within one epoch. Therefore, the complexity measures 
that quantify the generalization gap \cite{https://doi.org/10.48550/arxiv.1912.02178}
are inadequate and  
a different method of analysis is needed.

We will employ the measure, \emph{the effective dimension}, introduced in~\cite{complexity} to quantify the complexity of a computation performed by a neural network. 
In addition to providing a numerical estimate of the complexity of the computation, the measure introduced in~\cite{complexity} also characterizes the neural network processing at intermediate stages.
In this method one studies the intrinsic dimensionality of the internal feature representations, and calculates the effective number of independent features which the network uses
in order to distinguish between classes. 
We will evaluate a sequence of four effective dimensions that vary with the network
depth. Since a convolutional neural network operates by performing subsequent convolutions with small $3\times 3$ filters, as one goes deeper into the network, the feature representations become more and more nonlocal and hence transition from the UV to the IR. The resulting sequence of effective dimensions that we will calculate indicate how does the number of independent features vary as one passes from the UV scale to the IR scale.

The paper is organized as follows.
In section II we will construct the datasets of incompressible and weakly compressible chaotic and turbulent fluid
flows as well as various types of noise, 
and consider the inverse cascade turbulence scaling.
In section III we will discuss the neural network architecture and
the concept of effective dimension and use it to quantify the complexity of the task of
classification of real world images.
In section IV we will analyse the complexity 
of neural networks trained to distinguish turbulence fluid configurations from chaotic ones as well as varieties of artificial noise and some real world images. We will consider incompressible as well
as compressible fluid flows case and study the images of the vorticity configurations.
We will use an adversarial technique in order understand what features the neural network uses to distinguish chaos from turbulence.
Since, trained neural networks can show a good performance when tested on data coming from the same distribution as the training data, but a bad one when tested on other data,
we will study the test accuracy on chaos vs. turbulence data coming from simulations with different parameters or even different dynamics. 
Section V is devoted to a discussion and conclusions.
In appendix A we will present the analysis of the complexity of
fluid images representing a given component of the velocity vector field.

\section{Turbulence, Chaos and Noise}

To train the classifier, we generate snapshot datasets of turbulence, chaos and noise.
Turbulent and chaos datasets are generated for two-dimensional incompressible fluid flows as well as for 
compressible ones, where we extract the vorticity and velocity snapshots. The simulations are driven by a random force. Two noise datasets are generated, one noise dataset is generated by applying a Gaussian noise over an annulus in Fourier space which will also serve as the forcing of the fluid. The other is derived from the vorticity snapshot statistics. This section describes the simulation’s details and features as well as the noise generation process.

\subsection{Incompressible Fluid Flows}

The incompressible Navier-Stokes (NS) equations provide a mathematical formulation of the fluid flow evolution
at velocities much smaller than the speed of sound: 
\begin{equation}
\partial_t v^i + v^j\partial_j v^i =
-\partial^i p + \nu \partial_{jj} v^i - \alpha v^i + f^i,~~~~~~\partial_iv^i = 0  \ ,
\label{NS}
\end{equation}
where $v^i,i=1...d$  is the fluid velocity and  $p$ is the fluid pressure, $\nu$ is the kinematic viscosity, $\alpha$ is a linear frictional damping\footnote{This friction term would be responsible for removing energy at large scales, making the inverse energy cascade stationary \cite{2dTurbReview}.} and $f^i$ is an external
random force. In two space dimensions $(x,y)$, there is only one independent degree of freedom 
and one can rewrite the NS equations using the vorticity pseudoscalar $\omega=\epsilon_{ij} \partial^iv^j$:
\begin{align}
    \label{eq::vort}
    \partial_t \omega = (-1)^{p+1} \partial^{2p} \omega  - (v^i \partial_i)\omega + f_\omega \ ,
\end{align}

where $f_\omega = \epsilon_{ij}\partial^if^j$,
and we replaced for simulation reasons the second order viscous term in (\ref{NS})
by a hyperviscous term, in our case we set $p=2$ \cite{2dTurbReview}.

\begin{figure}[ht]
    \centering
    \includegraphics[width=0.47\textwidth]{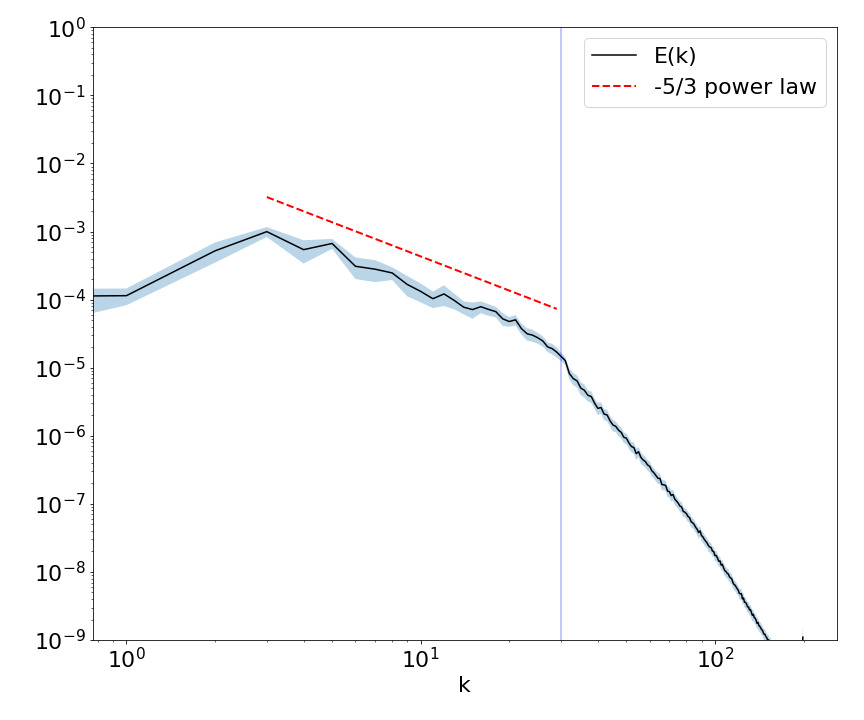}
    \includegraphics[width=0.47\textwidth]{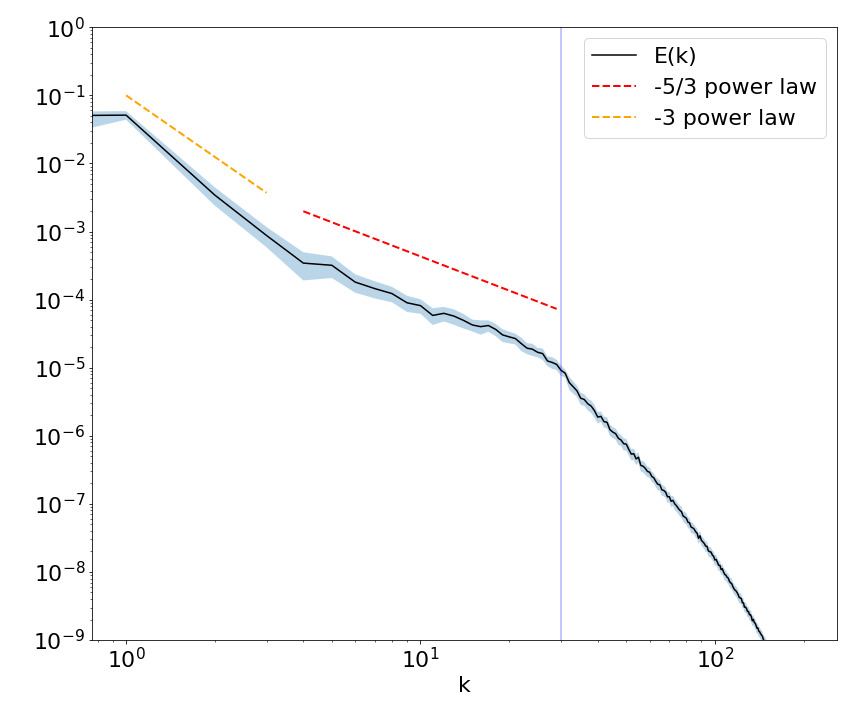}
    \includegraphics[width=0.47\textwidth]{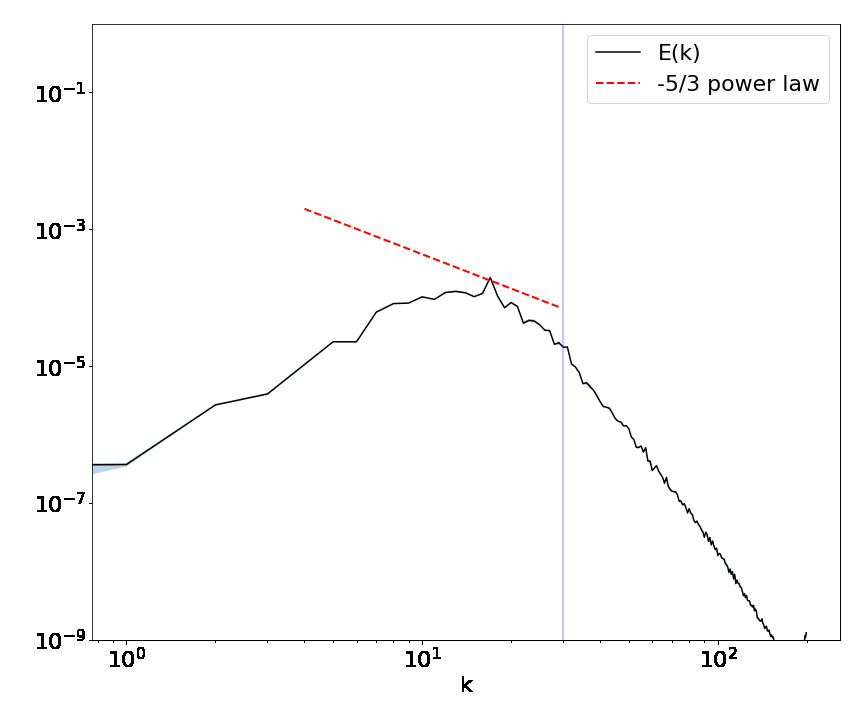}
    \caption{Energy spectrum as a function of the wave number of turbulent incompressible fluid in the inverse cascade. The vertical line
    is the forcing scale. On the top left one observes
    the $-5/3$ Kolomogorov scaling in the inertial range and  on the top right one sees evidence for the energy condensate through the $-3$ scaling. The bottom figure shows the energy spectrum of the fluid in the chaotic regime where we note the lack of scaling.}
    \label{fig:energySpec_incomp}
\end{figure}

\begin{figure}[ht]
\vspace*{-0.5cm}
\hskip -1cm
    \begin{subfigure}{0.24\textwidth}
        \includegraphics[width=1.25\textwidth]{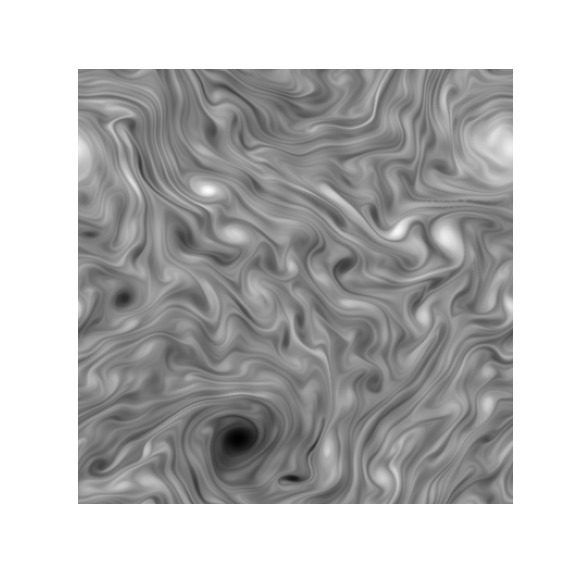}
        \vspace*{-1.5cm}
        \caption{}
        \label{fig:incomp_turb_1}
    \end{subfigure}
    \begin{subfigure}{0.24\textwidth}
        \includegraphics[width=1.25\textwidth]{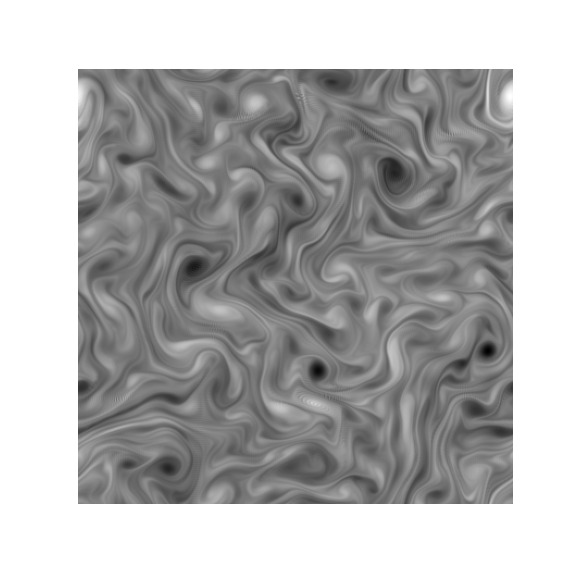}
        \vspace*{-1.5cm}
        \caption{}
        \label{fig:incomp_turb_2}
    \end{subfigure}
    \begin{subfigure}{0.24\textwidth}
        \includegraphics[width=1.25\textwidth]{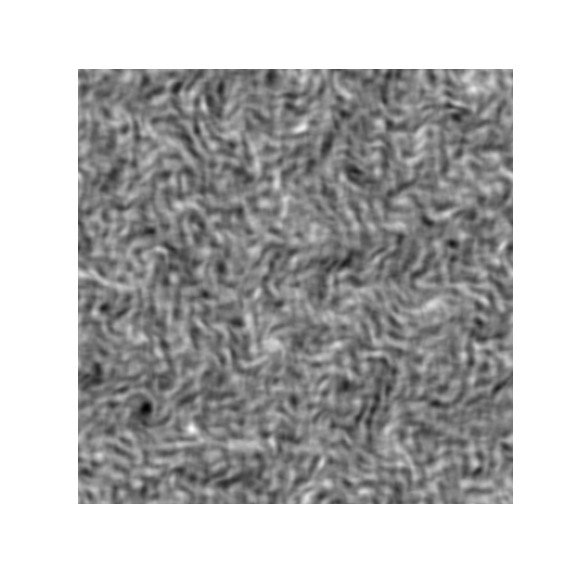}
        \vspace*{-1.5cm}
        \caption{}
        \label{fig:comp_turb_1}
    \end{subfigure}
    \begin{subfigure}{0.24\textwidth}
        \includegraphics[width=1.25\textwidth]{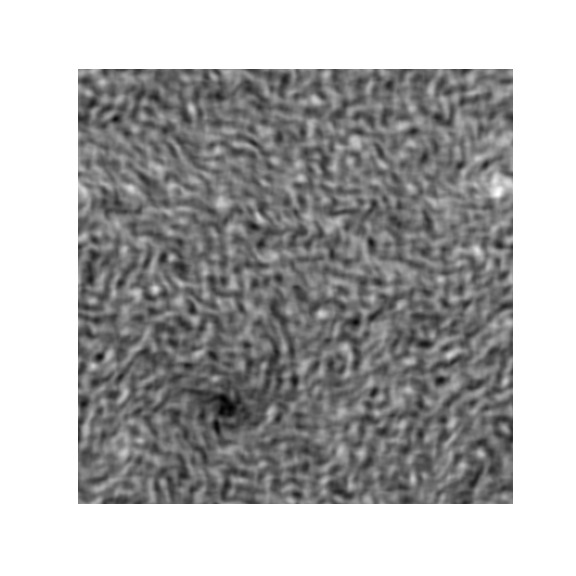}
        \vspace*{-1.5cm}
        \caption{}
        \label{fig:comp_turb_2}
    \end{subfigure}

    \hfill\\
\hskip -1cm
\begin{subfigure}{0.24\textwidth}
        \includegraphics[width=1.25\textwidth]{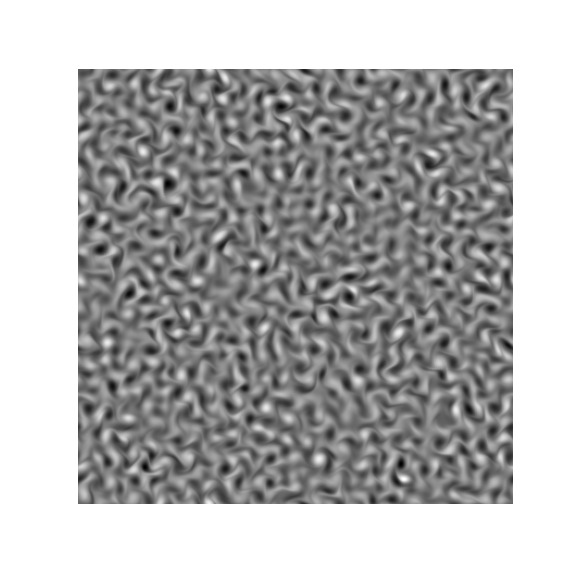}
        \vspace*{-1.5cm}
        \caption{}
    \end{subfigure}
    \begin{subfigure}{0.24\textwidth}
        \includegraphics[width=1.25\textwidth]{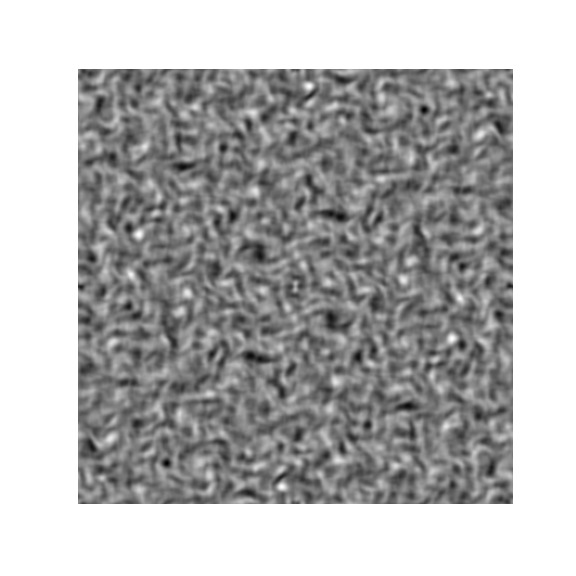}
        \vspace*{-1.5cm}
        \caption{}
    \end{subfigure}
    \begin{subfigure}{0.24\textwidth}
        \includegraphics[width=1.25\textwidth]{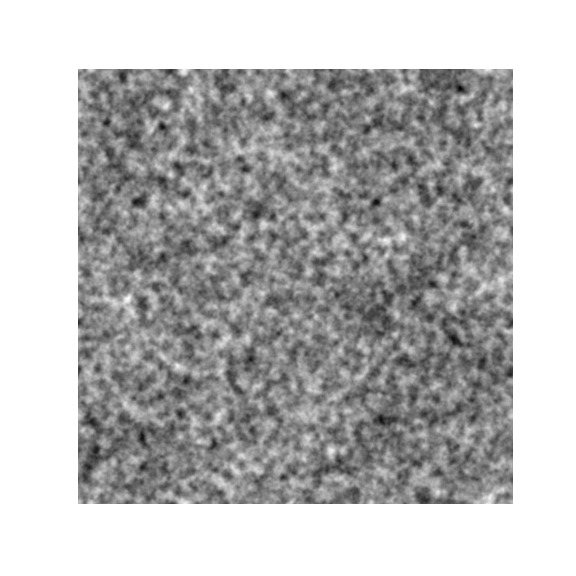}
        \vspace*{-1.5cm}
        \caption{}
    \end{subfigure}
    \begin{subfigure}{0.24\textwidth}
        \includegraphics[width=1.25\textwidth]{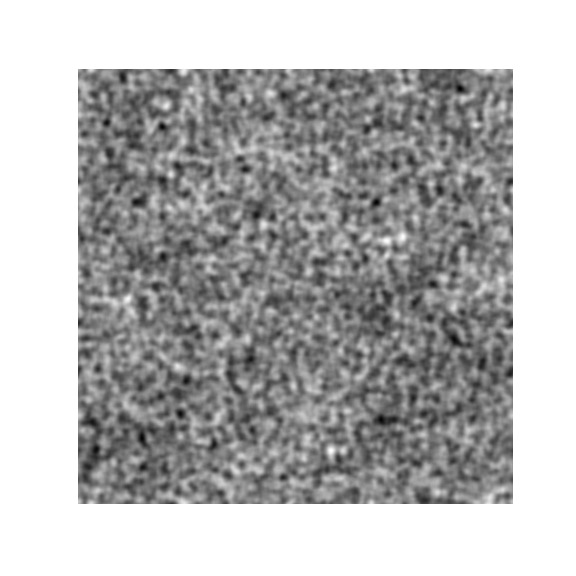}
        \vspace*{-1.5cm}
        \caption{}
    \end{subfigure}
    \caption{Sample images from the turbulent incompressible data~(a,b), turbulent weakly compressible data~(c,d), chaotic incompressible data~(e), chaotic weakly compressible data~(f), fourier noise from incompressible data~(g), fourier noise from weakly compressible data~(h).}
    \label{fig:samples}
\end{figure}

We use a divergence-free and statistically homogeneous and isotropic
Gaussian random forcing function, and
generate a dataset of two-dimensional incompressible fluid flows by numerically evolving the vorticity equation (\ref{eq::vort}).  The random forcing function has support in an annulus in Fourier space around $k_{forcing}$, and 
the system is numerically evolved with periodic boundary conditions with $v = (0,0)$ as initial condition on a uniform spatial grid. We use a dealiased spectral method code with Crank-Nicolson time stepping \cite{specmeth} and perform an ensemble of simulation with a resolution of $N^2 = 400^2$ and forcing $k_{forcing} \sim 15$. 
Due to the difficulty in including a linear friction term ($-\alpha \omega$) in the weakly compressible case in the next section, we avoid using a friction term in the incompressible case as well. The lack of friction can lead to energy condensates which, in physical space leads to the formation of two large vortex with opposite signs (see Fig.~\ref{fig:incomp_turb_1} and right  of Fig.~\ref{fig:incomp_turb_2}) \cite{2dturb,mcwilliams_1984,PhysRevLett.99.084501}. 
We isolate the turbulent regime of the simulation by taking snapshots, where we can observe the expected $-5/3$
K41 scaling of the energy (\ref{EK}) in the inverse cascade \cite{doi:10.1063/1.1762301,doi:10.1063/1.1692443}. We note that 2D turbulence has the possibility of producing a double cascade as shown in \cite{doi:10.1063/1.1762301} and numerically produced in \cite{PhysRevLett.81.2244,PhysRevE.82.016307} though we do not observe the direct cascade in our simulations. In figures \ref{fig:energySpec_incomp} we show the energy spectra of the turbulent snapshot (in this case we have $Re \sim 2100$). 
As noted, the lack of friction leads to an energy condensate which appears as a build up of energy at  low wavenumbers. We can see evidence of the energy condensate on the right of figures \ref{fig:energySpec_incomp} where there is a $-3$ power law. The chaotic dataset is defined 
by the set of snapshots in the time steps before the energy cascade is observed. 

\subsection{Weakly Compressible Fluid Flows}
\label{s.weakly}

We denote $x^{\mu} = (ct, x^i), i=1,2$ and $\eta_{\mu\nu} = diag[-1,1,1]$. In the following we will set $c=1$.
For the weakly compressible case, we consider a $(2+1)$-dimensional conformal perfect fluid with random forcing
$f^{\mu}$:
\begin{equation}
	\label{eq::rel_ns}
	\partial_{\mu} T^{\mu\nu} = f^{\nu} \ , 
\end{equation}
where the stress-energy tensor reads:
\begin{equation}
T^{\mu\nu} = \frac{\rho}{2}\left(\eta^{\mu\nu} + 3 u^{\mu} u^{\nu}\right)   \ . 
\end{equation}
$u^{\mu} = (\gamma, \gamma \beta^i), \beta^i = \frac{v^i}{c}, \gamma = \frac{1}{\sqrt{1-\beta^2}}, u_{\mu}u^{\mu}=-1$ is the 3-velocity vector field,
and we used the conformal equation of state relating the energy density $\rho$ to the
pressure $p$, $\rho = 2p$. 

For the purpose of evolving (\ref{eq::rel_ns}) numerically we set $f^0=0$ and write them
explicitly as:
\begin{eqnarray}
		\partial_t D + \partial_i S^i &=& -\nu \partial^4 D \ , \nonumber\\
		\partial_t S^i + \partial_j (S^j v^i+ \frac{\rho}{2} \delta^{ij}) &=& -\nu \partial^4 S^i + f^i \ ,
\end{eqnarray}
where $D = \frac{3}{2}\rho (\gamma^2-1)$, $S^{i} = \frac{3}{2}\rho \gamma^2 v^i$.

We then evolve the system numerically with periodic boundary conditions and $(\rho, v_i) = (1,0)$ as initial the condition on a uniform spatial grid. We use finite difference for the spatial derivatives and third order Runge-Kutta for time stepping. Dealising using the $2/3$ rule in the compressible case is numerically prohibitive so instead opt to include a fourth order hyperviscosity term \cite{Westernacher-Schneider:2015,Westernacher-Schneider:2017}. Similarly to the incompressible case, we perform an ensemble of simulations with a resolution of $N^2 = 400^2$ with forcing at $k_{forcing} \sim 15$. 

\begin{figure}[ht]
    \centering
    \includegraphics[width=0.47\textwidth]{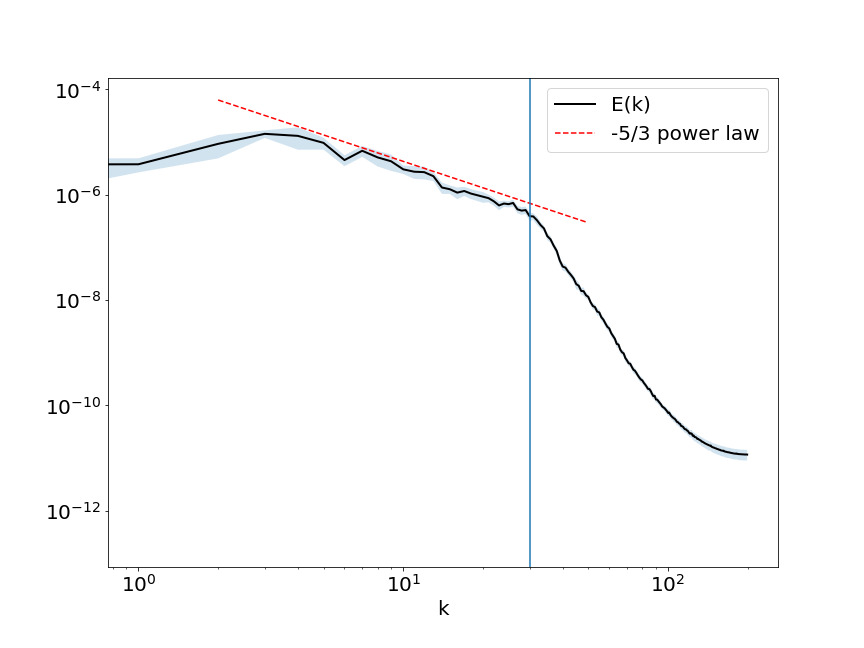}
    \includegraphics[width=0.47\textwidth]{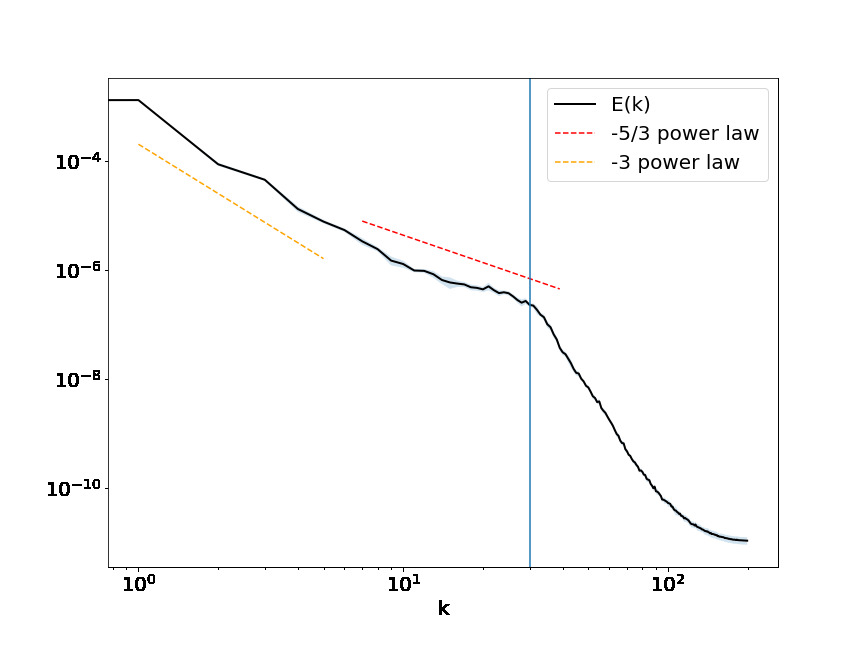} 
    \caption{Energy spectrum as a function of the wave number of turbulent weakly compressible fluid. The vertical line
    is the forcing scale. On the top left one observes
    the $-5/3$ Kolomogorov scaling in the inertial range and  on the top right one sees evidence for the energy condensate through the $-3$ scaling.}
    \label{fig:energySpec}
\end{figure}

As with the incompressible case, we isolate the turbulent regime of the simulation by taking snapshots where we can observe the $-5/3$ energy cascade. In figures \ref{fig:energySpec} we show the energy spectra of the turbulent snapshot. We also see evidence of the energy condensate as in the incompressible case in the right panel of figure \ref{fig:energySpec}. The chaotic dataset is the set of snapshots in the time steps before the energy cascade is observed.

\begin{figure}[ht]
    \centering
    \includegraphics[width=0.47\textwidth]{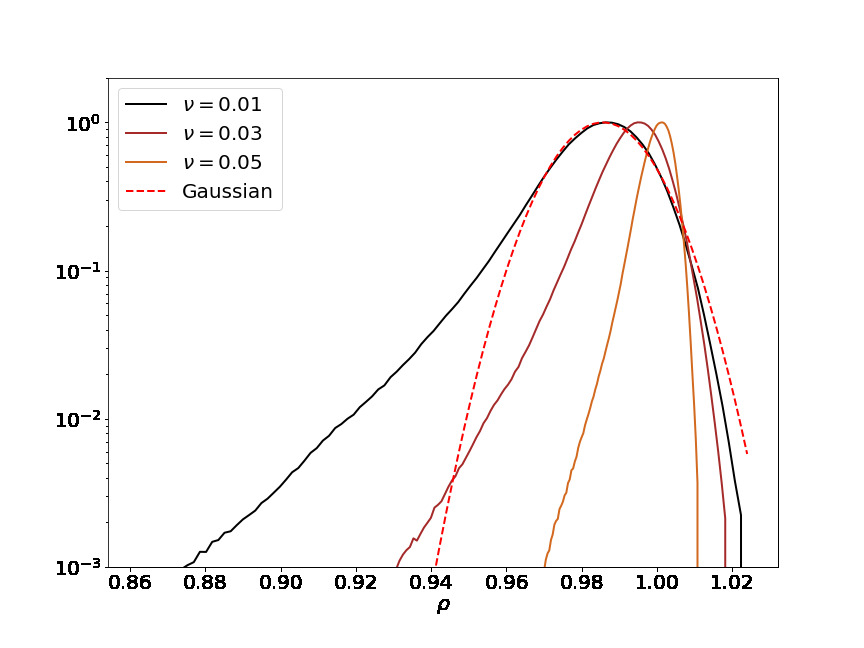}
    \includegraphics[width=0.47\textwidth]{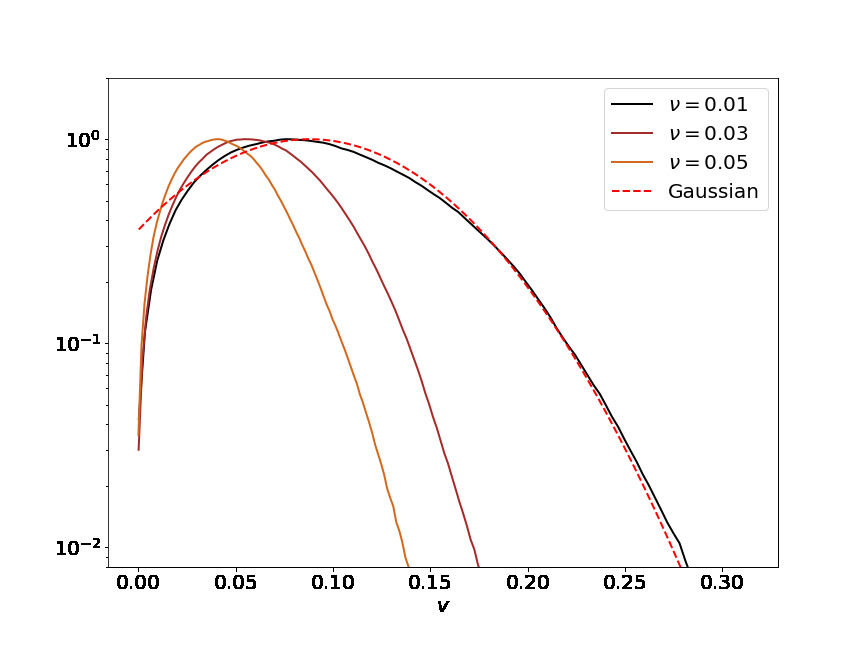}
    \caption{Probability distribution function for $\rho$ and $v$ from the weakly compressible simulations for $\nu=\{0.01,0.03,0.05\}$. We also display a Gaussian with mean and standard deviation fitted to the $\nu=0.01$ case.}
    \label{fig:DensPDF}
\end{figure}

Figure \ref{fig:DensPDF} shows the densities and velocities distribution of the final time step of some simulation to show evidence of the weak compressibility. We show these distributions for various $\nu$, the training dataset is constructed with $\nu=0.03$. 

\subsection{Noise}
\label{s.noise}

As mentioned previously, we generate two noise datasets. One is simply given by the same method we generate the streamfunction for the random forcing. We use the label ``noise'' to denote this dataset. The second noise dataset is generated by first computing the means and standard deviations of Fourier transform coefficients of 1000 random turbulence images. Then coefficients are sampled from a corresponding Gaussian distribution and a inverse Fourier transform is performed. The label ``noise fourier'' is used for this noise dataset.

The output datasets are black and white images of the vorticity, velocities and noise. The images are generated using python's matplotlib package \cite{matplot} with interpolation and with a pixel resolution of $576 \time 576$ with white boundaries which are cropped before entering the NN. 
 
\section{Neural Network Complexity}

\subsection{Deep Convolutional Neural Network}

A typical deep convolutional neural network takes as input an RGB image ($H \times W \times 3$), performs convolutions with a set of $C$ small $3\times 3$ filters, acts on the outcome with a ReLU nonlinearity\footnote{$ReLU(x)=x$ if $x\geq 0$ and $ReLU(x)=0$ if $x<0$.} and repeats the process many times\footnote{Here we omit mentioning batch normalization layers and residual/skip connections which are not essential for our discussion.}, occasionally reducing the image size and increasing the number of channels $C$. At the end, a final linear layer and a sigmoid will yield the probability that the image belongs to a given class\footnote{For simplicity we consider here a classification task with one positive and one negative class.}.
\begin{figure}[h]
    \centering
    \includegraphics[width=\textwidth]{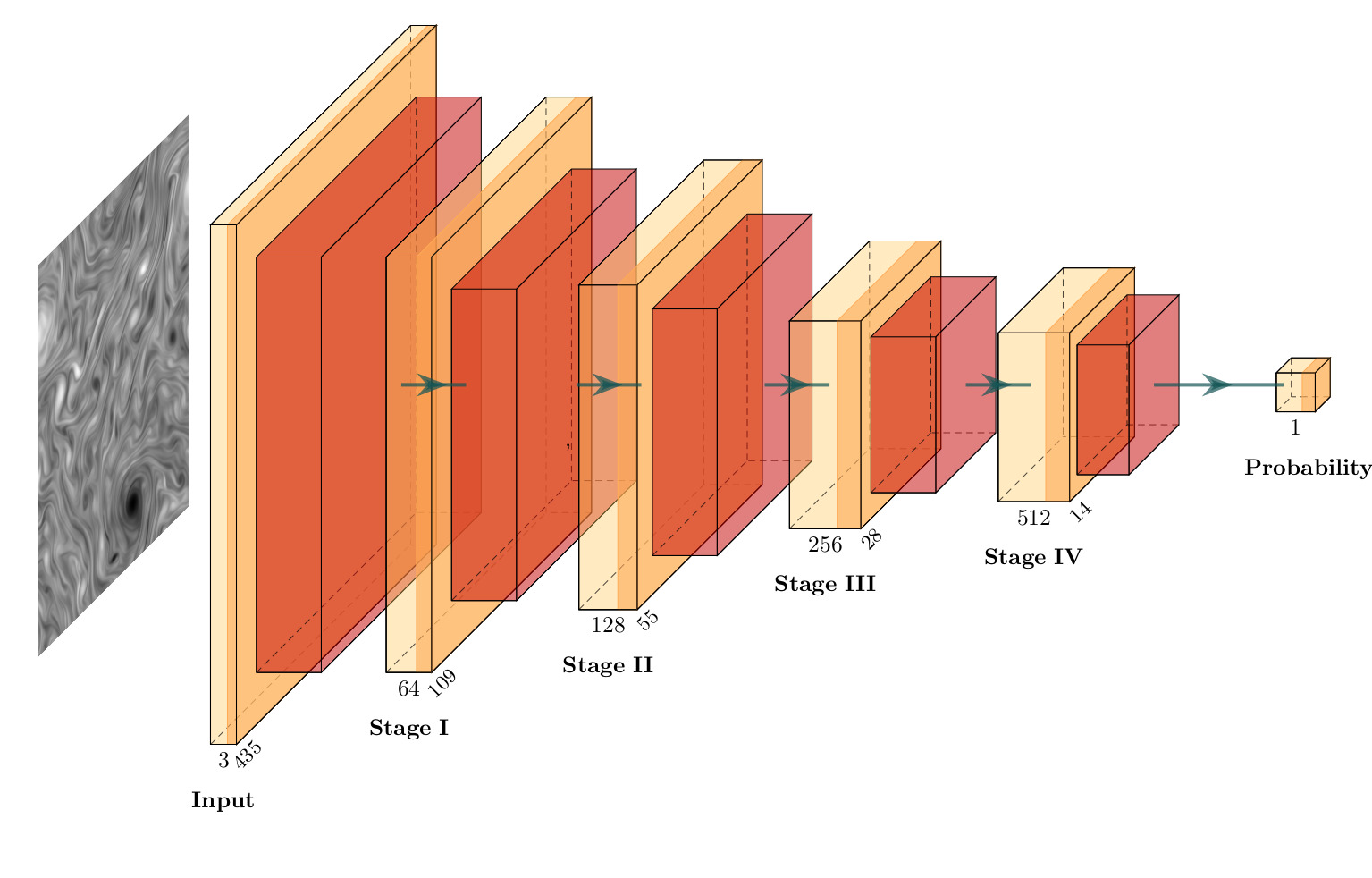}
    \caption{Schematic of the CNN. Each stage represents a set of convolutional layers.}
    \label{stages}
\end{figure}
All images considered in the present paper are grayscale, so the RGB components are always set to be identical.

During the processing of an image by the neural network as sketched in Fig.~\ref{stages}, the original pixels are transformed to successively lower resolution images but with a much higher number of channels. These intermediate images are often referred to as  internal feature representations of the network and the number of channels is defined to be its dimensionality (here $64$, $128$, $256$ and $512$).

Since the processing in each ``stage'' is a succession of convolutions with a small filter, we see that the internal feature representations are more and more nonlocal as we go deeper into the network. Due to the small filter size, the nonlocality increases only gradually.
We should note, however, that because of the multiple ReLU nonlinearities, the ``stages'' cannot be thought of just as straightforward coarse-grainings but rather as some intrinsically nonlinear transforms.

A final important point is that a neural network used for classifying images into two classes, aims to construct (during training) the internal feature representations so that they would be optimal for distinguishing between the two classes\footnote{The final representation should be such that the two classes would be linearly separable.}. Hence, studying the internal representations e.g. in a noise versus turbulence task, we aim to quantify at various levels of nonlocality  how many of the features are necessary for the neural network to distinguish the two classes.  This can be precisely defined through the notion of \textit{effective dimension} introduced in~\cite{complexity}.

\subsection{Effective Dimension or Linear Complexity}
\label{s.lincomplexity}

The paper \cite{complexity} introduced two complementary quantitative notions which characterize the difficulty of a given calculational task as performed by a neural network. The first one, \emph{complexity}, characterizes to what extent nonlinearity (i.e. turning neurons on and off by the ReLU nonlinearity) is necessary for the given computation. It is defined as a Shannon entropy of binary ``nonlinearity'' variables, and although calculable, is quite computationally expensive to compute. A second complementary notion, \emph{the effective dimension}, analyzes instead the intrinsic dimensionality of the internal feature representations, and thus captures to what extent the particular internal features are linearly independent or not. It provides a simple numerical estimate of the effective number of independent features. One can view it thus as a kind of \emph{linear complexity}.   
It is relatively efficient to compute, and also directly answers the question of interest for us, namely the counting of independent features employed by the network for distinguishing between two classes (e.g. turbulence and chaotic). 

The counting of independent features can be done individually for a given internal feature representation at a particular depth. In this paper, we will do it at the output of stages 1-4 as shown in Fig.~\ref{stages}
and obtain a sequence of four effective dimensions varying with depth. A convolutional neural network operates by performing subsequent convolutions with small $3\times 3$ filters, so as one goes deeper into the network, the feature representations become more nonlocal and hence transition from the UV to the IR. The resulting sequence of effective dimensions thus indicates how does the number of independent features vary as one passes from the UV scale to the IR scale.
For completeness, we review below the definition of effective dimension from~\cite{complexity}.

As we want to determine to what extent the different dimensions (i.e. channels) of the internal feature representation are independent or not, we accumulate the representations for a test set of the problem on which the network was trained\footnote{It is standard to separate the dataset into disjoint training part and a test set, and use the latter for determining accuracy and performing various analysis. }. Since we are interested in the independence of the channels understood as individual features, we reorder the resulting data for $N$ images ignoring spatial positions:
\eq
N \times H \times W \times C \longrightarrow (N\cdot H \cdot W) \times C \ .
\eqx
We then perform Principal Component Analysis (PCA) of the resulting data and obtain the explained variance ratio $r_i$ of each component. Since by definition the explained variance ratio sums up to 1
\begin{equation}
\sum_i r_i = 1     \ .
\end{equation}
we can formally compute its Shannon entropy $S[\{r_i\}]$. We then exponentiate it to define the effective dimension:
\begin{equation}
\label{e.effdim}
\text{effective dimension} = \exp \left(S[\{r_i\}]\right) \equiv \exp \left( -\sum_i r_i \log r_i \right) \ .
\end{equation}
This definition gives the expected answer $K$ for simple cases like
\begin{equation}
r_i = \f{1}{K} \quad \text{for}\; i\leq K \quad\quad r_i=0 \quad \text{for}\; i>K
\end{equation}
and smoothly extrapolates to other configurations. One of the advantages of the definition (\ref{e.effdim}) is that it does not require any additional parameter (like a cutoff for the explained variance ratio) to estimate dimensionality. It is also readily calculable as it uses PCA.

\subsection{Effective dimension for standard image classification tasks}

In ref.~\cite{complexity}, the effective dimension was computed for a range of image classification tasks including colored versions of MNIST (a dataset of hand written digits), KMNIST (a dataset of Japanese script characters), FashionMNIST (a dataset of various types of clothing items) and CIFAR-10 (a dataset of color images belonging to 10 classes). 
For all cases, the same network architecture was used -- a ResNet-56 network for small $32\times 32$ images -- with an identical training protocol.
The main characteristics of the results obtained in~\cite{complexity} were:
\begin{enumerate}
\item The effective dimensions summed over the layers increased with the intuitive complexity of the classification task, namely
\eq
\text{MNIST} < \text{KMNIST} < \text{FashionMNIST} < \text{CIFAR-10}
\eqx
\item For all these classical image datasets, the effective dimension for individual layers\footnote{More precisely after each residual block of the ResNet-56 network comprising two convolutional layers.} \emph{increased} as one went deeper into the neural network with a decrease only just at the very end before the final classification layer.
\end{enumerate}

The former result is a qualitative validation of the use of effective dimension as a measure of task complexity, with the more difficult classification tasks requiring that the neural network has to construct richer (more independent) feature representations at intermediate stages of processing. 

The latter result indicates that more and more nonlocal features are needed for performing the classification. Indeed the differentiation between \emph{a car} and \emph{a truck} (two example classes from CIFAR-10) requires piecing together local information (presence of wheels, metalic finish etc.) and basing the decision on a global level, hence in the deeper layers of the network. These layers, therefore, have to represent the various relevant global contexts leading to a higher effective dimension of the feature representations deeper in the neural network. 

One can interpret the above behaviour in the Physics language, by stating that there is crucial information in the IR and the UV is not sufficient. That being said, one has to remember that the neural network feature representations are not linear and one expects that the neural network constructs more and more nontrivial composite features. Moreover, the features at the deeper level can contain not only purely IR characteristics but also e.g. some integrated UV features if they would be necessary for performing the classification.

The decrease of the effective dimension at the very end of the network is in fact quite natural as the images have to be classified into a small number of classes (10 in the examples considered in~\cite{complexity}).
So one can interpret this decrease as integrating the information into composite features directly necessary for performing the final classification, and discarding irrelevant data.

\section{Turbulence Classifiers}

Our main goal in this paper is to analyze turbulence from the perspective of deep neural networks. We will adopt the techniques reviewed in the previous section for analyzing neural networks which are trained to distinguish turbulence fluid configurations from chaotic ones as well as varieties of artificial noise and some real world images. Apart from the standard incompressible case, we also consider simulations of the weakly compressible variant (see section~\ref{s.weakly}). We primarily analyze vorticity configurations but also consider images representing a given component of the velocity vector field in appendix~\ref{app:vel}.
The key questions that we address are:
\begin{enumerate}
\item What is the relative complexity of the various classification tasks involving turbulence?
\item How does the pattern of complexity, as measured by the effective dimension, change with depth as we go inside the neural network? How does it compare with classifying real world images?
\item Can we understand what features the neural network uses to distinguish chaos from turbulence?
\end{enumerate}

After a brief description of the technical setup, in section~\ref{s.catsdogs} we analyze examples with a dataset of cat and dog images. This serves two purposes. First, since fluid simulation profiles are clearly quite different from real-world images we can check whether our methodology of assessing complexity as a function of scale works as expected. Second, the pattern of complexity (effective dimension) for distinguishing real-world images (cats vs. dogs) will be a point of reference for the tasks of distinguishing turbulence from chaos and various synthetic ``pseudo-flow'' datasets which we discuss in section~\ref{s.effdimchaosnoise}. We then proceed to identify the key features used by the neural network to distinguish turbulence from chaos in section~\ref{s.adversarial}, and discuss the performance of a trained neural network on test data coming from different physical simulations in section~\ref{s.othertest}.

\subsection{Technical setup}

For the numerical experiments performed in the present paper we used a standard {\tt resnet-18} neural network as used for \textit{Imagenet} \cite{ILSVRC15}. All input images were provided at a resolution of $435\times 435$ pixels. Prior to input to the neural network, each image was normalized to have zero mean and unit standard deviation. Since vorticity configurations can be naturally represented as a grayscale image, the images of cats and dogs were also transformed to grayscale. The {\tt resnet-18} network was, however, unmodified and the grayscale images were represented as RGB images with identical RGB components.

From each class, we set aside 1000 images for evaluating accuracy and effective dimensions. Training was performed with an Adams optimizer with a small weight decay 1e-4 added, in order to reduce the variance of the obtained effective dimensions. Training was performed until test accuracy of 99\% was reached\footnote{Or a cut-off of 20 epochs}. For most cases this occurred already after 1 epoch. 
We repeated each experiment 5 times starting from different random initializations of the neural network.

\subsection{Effective dimensions -- turbulence vs. real world images}
\label{s.catsdogs}

\begin{figure}[t]
    \includegraphics[width=0.45\textwidth]{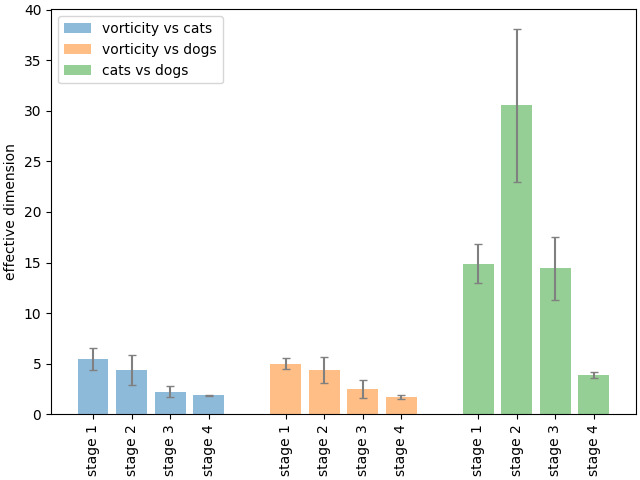}
    \includegraphics[width=0.45\textwidth]{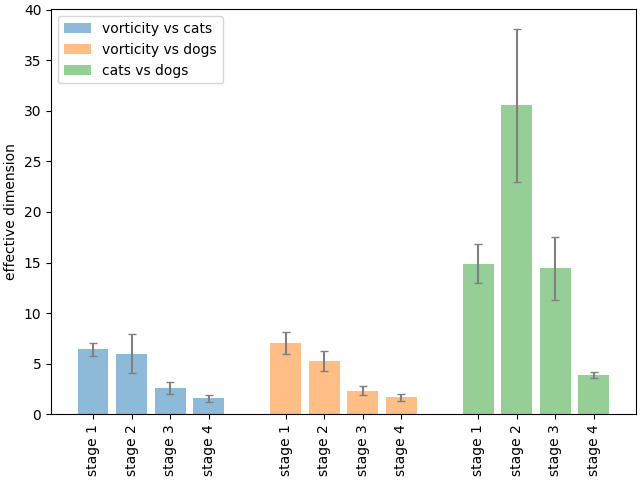}
    \caption{The left panel shows the effective dimensions for images of weakly compressible turbulence vorticity vs. cats and dogs as well as for classifying between cats and dogs. The right panel shows the incompressible case.}
    \label{fig:effdimspets}
\end{figure}

As we would like to compare the complexity of differentiating real world images versus differentiating turbulence from chaos, we made a series of experiments involving images of cats and dogs\footnote{Downloaded from \url{https://www.microsoft.com/en-us/download/details.aspx?id=54765}.}. 
9935 images were selected from each class, which is similar to the number of images of turbulence vorticity.
The images were resized to 435x435 (ignoring aspect ratio) and changed to grayscale. 

As a point of reference, we compute the effective dimensions for classifying cats vs. dogs using the above data.  
In Fig.~\ref{fig:effdimspets} (green bars on the right) we show the effective dimensions computed after 25 epochs averaged over 5 random initializations. The obtained accuracy was\footnote{The target accuracy of $99\%$ used for the remaining experiments in this paper could not be reached in this case, hence we chose a fixed number of epochs here.} $92.2 \pm 1\%$. 
In Fig.~\ref{fig:effdimspets} we also show, as a toy example, effective dimensions for distinguishing turbulence vorticity from cats and dogs. 

Let us again emphasize that the latter task of distinguishing real world images from turbulence has no intrinsic interest, but serves only as a cross-check of our methodology and the observables that we use to probe the network. Indeed in this case we have very clear expectations of what should be the qualitative behaviour of the neural network with regard to complexity of the task. We can thus check whether this is reflected in the profiles of effective dimensions.

Several comments are in order here.
First, the effective dimensions for cats vs. dogs strongly increase as we go into the network, with stage~3 being not less than stage~1. This qualitative picture thus confirms the findings of ref~\cite{complexity} in this respect. 
The interpretation is that for the classification of cats vs dogs, more global (IR) and non-local features are used, with high values even for stage~3.
The fall-off towards the end of the network is associated with approaching the classification layer with just two classes. 

Second, the problem of distinguishing turbulence from cats or dogs has a quite different profile of complexity vs. depth. The maximal value is obtained for stage~1 and then steadily decreases, even though the number of channels \emph{increases} as we go deeper into the neural network. This indicates that UV local features are most important for distinguishing turbulence vorticity from images of cats and dogs. Clearly, this is true as even a small local patch of a vorticity profile is at first glance completely different from an analogous patch of a real-world image.

Third, the total sum of the effective dimensions is \emph{significantly greater} for cats vs. dogs than for turbulence vs. cats or dogs. This sum can be considered as a measure of the overall complexity of the problem, and clearly distinguishing cats from dogs is more difficult than distinguishing turbulence from either. 

The latter two points serve primarily as a consistency check of our methodology, while the profile of effective dimensions for cats vs. dogs will be a point of reference for the physically interesting classification problems of distinguishing turbulence from chaos or various kinds of noise, which we will discuss in the following section.

\subsection{Effective dimensions -- turbulence vs. chaos or noise}
\label{s.effdimchaosnoise}

Let us first consider a neural network which distinguishes the vorticity profiles of incompressible turbulence versus (incompressible) chaos. The network reaches 100\% accuracy already after a single epoch. The resulting effective dimensions are:

\bigskip

\begin{tabular}{c|c|c|c|c}
weights & stage 1 & stage 2  & stage 3  & stage 4 \\
\hline
incompressible turbulence vs. chaos & $7.5 \pm 3.4$ & $9.3 \pm 6.8$ & $3.2 \pm 1.6$ & $1.6 \pm 0.2$
\end{tabular}

\bigskip

\noindent{}where we indicated the average and standard deviation across 5 repetitions with different random initializations of the network. We see that the profile of the effective dimensions is definitely more spread towards the UV than for the cats vs dogs case. In particular stage~3 is significantly less than stage~1.

In addition, we see that the absolute numbers are also much smaller, which indicates that distinguishing turbulence from chaos is a much easier problem for the neural network than distinguishing real-world images.
Let us emphasize, that this does not mean that the turbulence and chaos vorticity images are too simplistic by themselves. Indeed, if we evaluate the test set on randomly initialized networks, or on the same network but trained on \textit{ImageNet}, i.e. on a completely different task we obtain much larger effective dimensions:

\bigskip

\begin{tabular}{c|c|c|c|c}
weights & stage 1 & stage 2  & stage 3  & stage 4 \\
\hline
imagenet & 24.4 & 51.3 & 91.3 & 42.1 \\
random initializations & $18.3 \pm 3.4$ & $36 \pm 9.9$ & $50 \pm 7.8$ & $44.4 \pm 9.1$
\end{tabular}

\bigskip

\noindent{}The above indicates, that the pixel structure of the vorticity profiles is not really trivial, as random convolutional filters and nonlinearities can generate relatively high dimensional representations at various scales (here depths of the network).
On the other hand, the very much smaller values appearing for the trained neural networks indicate, that one can discard a huge part of this specific information in order to distinguish turbulence from chaos. We will return to the question of what type of information is used in this case in the following section.  

\begin{figure}[t]
    \centering
    \includegraphics[width=0.47\textwidth]{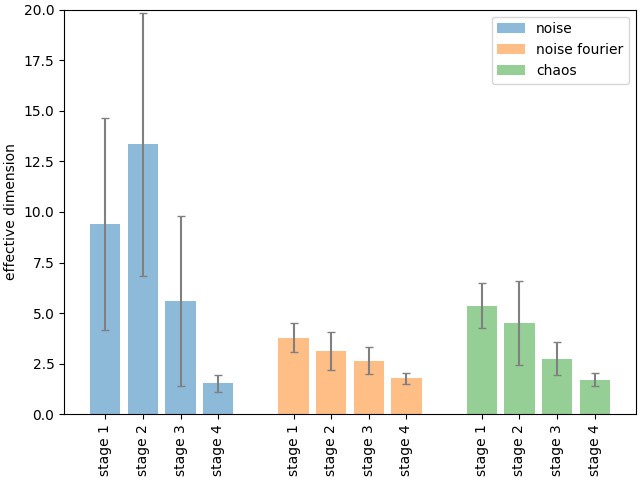}
    \includegraphics[width=0.47\textwidth]{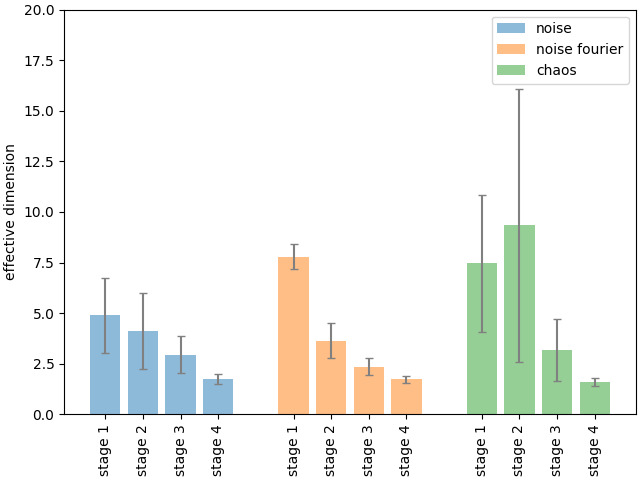}
    \caption{Effective dimensions for classifying weakly compressible turbulence vorticity (left) and 
    incompressible turbulence vorticity (right).}
    \label{fig:vorticity}
\end{figure}

In Fig.~\ref{fig:vorticity}, we show the pattern of effective dimensions for classifying turbulence vorticity images from chaos as well as the two types of noise discussed in section~\ref{s.noise}. Of these noise\_fourier is especially interesting, as it encodes just the magnitudes of the Fourier coefficients of turbulence vorticity ignoring any correlations or higher order nonlinear effects. We show results both for the weakly compressible case as well as for the incompressible case.

There are several points to note here.
First, one can observe that all these problems are relatively easy -- the effective dimensions are much smaller than for the cats vs. dogs case.
Second, the classification of turbulence vs noise is concentrated in the UV with a decreasing pattern of complexity vs. depth. The only cases which seem to require more processing at larger scales is the case of incompressible chaos vs. turbulence (and noise in the weakly compressible case). But even here, the increase only involves stage~2, while stage~3 is already significantly lower than stage~1, which is quite opposite to the case of real world images from Fig.~\ref{fig:effdimspets}.

\begin{figure}[t]
    \centering
    \includegraphics[width=0.47\textwidth]{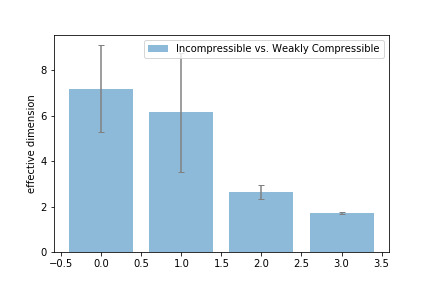}
    \caption{Effective dimensions for distinguishing incompressible from weakly compressible turbulence.}
    \label{fig:vorticity_incomp_comp}
\end{figure}

For completeness, we also checked the profile of effective dimensions when distinguishing weakly compressible turbulence from the incompressible one.
Somewhat surprisingly, we find that this problem is generally quite easy with rather low effective dimensions and again concentrated in the UV. The results are shown in Fig.~\ref{fig:vorticity_incomp_comp}. Looking at the sample images of incompressible and weakly compressible  turbulence vorticity (compare Fig.~\ref{fig:samples}a,b with Fig.~\ref{fig:samples}c,d), we see that indeed it is easy to distinguish the two ``by eye'', with the main difference being in the ``sharpness'' of the local structure. Hence this is compatible with the outcome of the effective dimension computation.

\subsection{Adversarial examples}
\label{s.adversarial}

In this section, we would like to focus on identifying what kind of information is used by the neural network when distinguishing turbulence vorticity images from the ones in the chaotic phase. The problem of identifying the reasons of a deep neural network decision is, in principle, a quite difficult one \cite{https://doi.org/10.48550/arxiv.1702.08608,Ribeiro2016WhySI,lrp}. This is even more so for the kind of images that we are dealing with now, namely vorticity profiles of turbulence or chaos. For real world images, various methods of interpretability or explainability typically serve to identify the pixels in the input image which were most important for performing the classification (see e.g. \cite{https://doi.org/10.48550/arxiv.1412.6572,lrp}). Such an approach is largely inapplicable to the images of turbulent or chaotic flow, as most images are spatially homogeneous\footnote{Turbulent images with two large vortices are an exception. This feature occurs, however, only in a subset of turbulent images and hence cannot be used as a key reason for classifying an image as turbulence.} and we expect the classification to arise from differences in local statistics or correlation structure, neither of which is localized in some specific subregion of the image.

In order to overcome the above difficulties, we adopted a procedure loosely motivated by the notion of adversarial examples. We evaluate a neural network trained to distinguish turbulence from chaos on a set of artificial examples which capture only a subset of the physical characteristics of either turbulence or chaos. We check whether the neural network still classifies these proxy images as turbulence or chaos. If this remains so, it indicates that the physical characteristics which were dropped in constructing the proxy images are not essential for the classification.

In addition, we evaluate the network which was trained on incompressible turbulence also on weakly compressible data and vice-versa. The results are shown in tables~\ref{Tab:incomp_attack_final} and~\ref{Tab:comp_attack_final}. These results are presented for a sample neural network for which the interpretation may be most clear.

\begin{table}[ht]
\begin{tabular}{l|l|l|}
\cline{2-3}
                                    & Chaos & Turbulence \\ \hline
\multicolumn{1}{|l|}{Compressible Chaos}         &   100\%    &       0\%     \\ \hline
\multicolumn{1}{|l|}{Compressible Turbulence}    &    0\%   &       100\%     \\ \hline
\multicolumn{1}{|l|}{Incompressible Turbulence with $k_{forcing} \sim 7$}    &    0\%   &       100\%     \\ \hline
\multicolumn{1}{|l|}{Compressible Noise Fourier} &     0\%  &        100\%    \\ \hline
\multicolumn{1}{|l|}{Incompressible Noise Fourier} &   0\%    &      100\%      \\ \hline
\multicolumn{1}{|l|}{Compressible Chaos Noise Fourier} &     100\%  &        0\%    \\ \hline
\multicolumn{1}{|l|}{Incompressible Chaos Noise Fourier} &   100\%    &      0\%      \\ \hline
\end{tabular}
\caption{Output classification result of adversarial inputs into the incompressible vorticity vs. chaos classifier. }
\label{Tab:incomp_attack_final}
\end{table}

\begin{table}[ht]
\begin{tabular}{l|l|l|}
\cline{2-3}
                                    & Chaos & Turbulence \\ \hline
\multicolumn{1}{|l|}{Incompressible Chaos}         &    100\%   &   0\%         \\ \hline
\multicolumn{1}{|l|}{Incompressible Turbulence}    &   0\%    &     100\%       \\ \hline
\multicolumn{1}{|l|}{Compressible Noise Fourier} &   0\%    &      100\%      \\ \hline
\multicolumn{1}{|l|}{Incompressible Noise Fourier} &    0\%   &     100\%       \\ \hline
\multicolumn{1}{|l|}{Compressible Chaos Noise Fourier} &     100\%  &        0\%    \\ \hline
\multicolumn{1}{|l|}{Incompressible Chaos Noise Fourier} &   100\%    &      0\%      \\ \hline
\end{tabular}
\caption{Output classification result of adversarial inputs into the weakly compressible vorticity vs. chaos classifier.}
\label{Tab:comp_attack_final}
\end{table}

The key datasets for our arguments are the Noise Fourier ones discussed in section~\ref{s.noise}.
They are constructed in the following way. The Fourier coefficients of all the images of interest (e.g. incompressible turbulence) are computed and their mean and standard deviation over the whole dataset is evaluated. Then images in the Noise Fourier dataset are constructed by random sampling of each Fourier coefficient from the appropriate Gaussian distribution then taking the inverse Fourier transform.
This procedure leads to a dataset which has the same Fourier spectrum as e.g. turbulence, but ignores any correlations between the Fourier coefficients, as well as neglects any nonlinear effects.
The relevant images for Noise Fourier both for turbulence and for chaos are classified by the neural network with high confidence respectively as turbulence and chaos both for incompressible and for weakly compressible case. This strongly indicates that the nonlinear correlations are not necessary for distinguishing turbulence from chaos.

\begin{figure}[ht]
    \centering
    \includegraphics[width=0.47\textwidth]{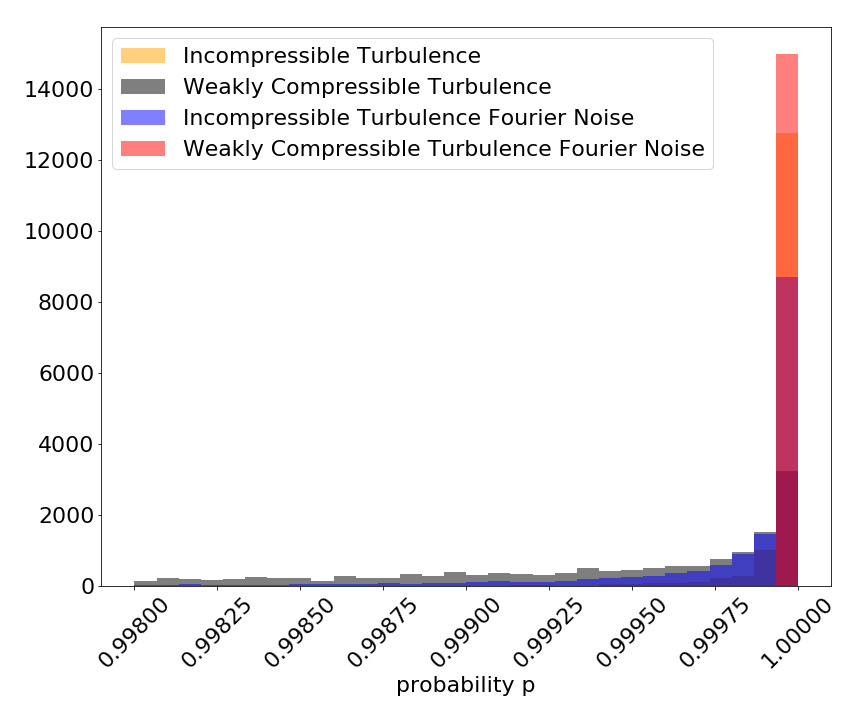}
    \includegraphics[width=0.47\textwidth]{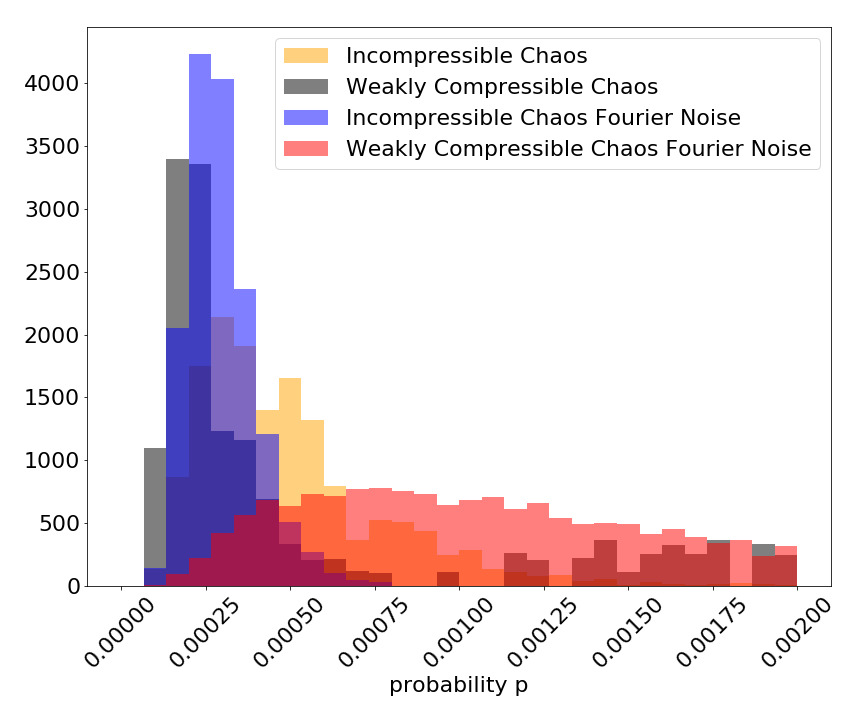} 
    \caption{Histograms of incompressible chaos vs. turbulence classifier probabilities evaluated on various test datasets. 
    $p=1$ indicates 100\% confidence that the given image is turbulence, while $p=0$ indicates 100\% confidence that the given image is chaos.
    Turbulence related data on the left, chaos related data on the right.}
    \label{fig:classifier_probs}
\end{figure}

Indeed, in Fig.~\ref{fig:classifier_probs}, we show histograms of the probabilities for various test sets. Basically all values are very close either to 0 or to 1, indicating that the neural network classifies these images with very high confidence either as turbulence or as chaos. So the lack of nonlinear correlations in the Turbulence Noise Fourier dataset is not reflected in a noticeable way in the neural network outputs.

\begin{figure}[ht]
    \centering
    \includegraphics[width=0.47\textwidth]{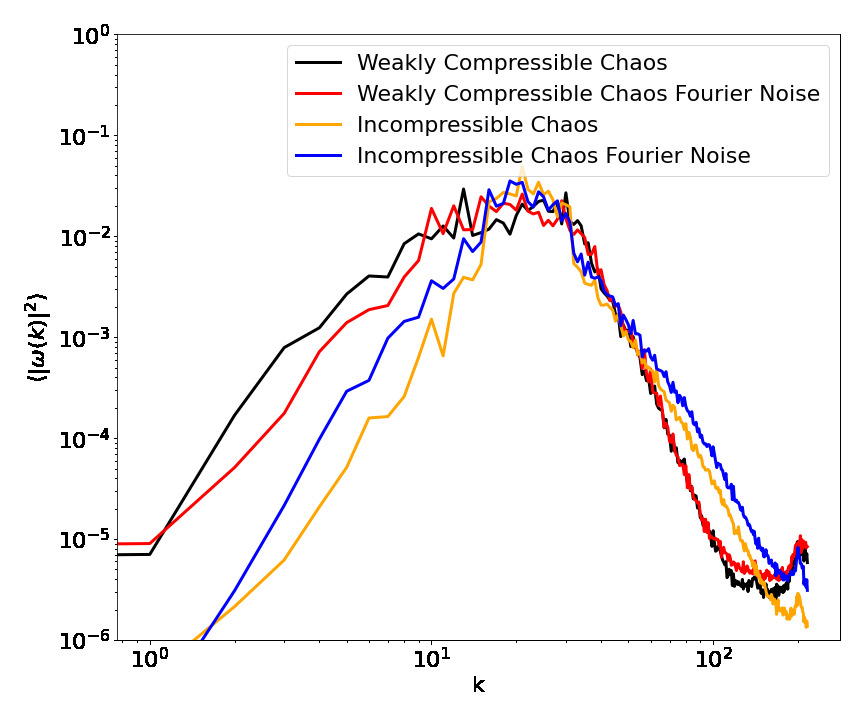}
    \includegraphics[width=0.47\textwidth]{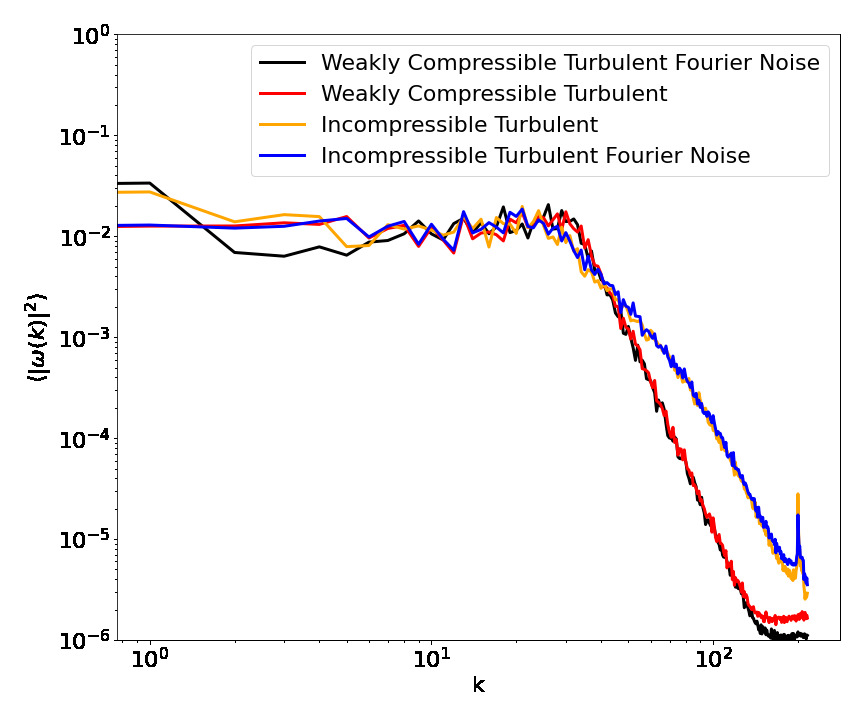} 
    \caption{$\langle |\omega (k)|^2 \rangle$ for sample normalized images used in the adversarial attacks. The left panel contains the chaotic data and its derived Fourier noise while the right panel contains the turbulent data and its derived Fourier noise. }
    \label{fig:2momentSpec_final}
\end{figure}

The above results are in agreement with the intuition that a neural network focuses on the simplest and most obvious differences which are enough for performing the classification. Indeed, as shown in Fig.~\ref{fig:2momentSpec_final}, the two point correlation spectra for the chaotic and turbulent vorticity are clearly different, and suffice to make the distinction. 
We may note, that the biggest difference between chaos and turbulence in Fig.~\ref{fig:2momentSpec_final} seems to be in the IR. This is consistent with the relatively high effective dimension at stage 2 for the turbulence-chaos classifiers in Fig.~\ref{fig:vorticity}.

Let us make a further comment on the relation of the effective dimensions for the incompressible turbulence vs chaos classifier shown in Fig.~\ref{fig:vorticity}(right) and the claim of the present section, that the key characteristic used for classification is essentially the Fourier spectrum, as shown in Fig.~\ref{fig:2momentSpec_final}. The effective dimensions roughly correspond to the number of independent \emph{scalar} features which are used at a given depth of the network\footnote{These features are evaluated in the same way at each point of the appropriately coarse-grained image~(see~(\ref{stages})).}. From Fig.~\ref{fig:vorticity}(right) we see that these are small single digit numbers. On the other hand, the Fourier spectrum is a function of the wave-vector $k$, and hence contains effectively many (scalar) degrees of freedom. The low effective dimensions indicate that the network uses only some simple composite information out of the Fourier spectrum and not its exact precise shape.

The fact that the incompressible classifier works correctly in the weakly compressible case and vice-versa can also be explained by the hypothesis that the two point spectra are the key features taken into account by the neural network. 

As a last check, we also evaluated the incompressible classifier on turbulent vorticity images generated with a forcing scale $k_{forcing} \sim 7$, \emph{different} from the forcing scale at which the neural network was trained ($k_{forcing} \sim 15$). 
The fact that these images are still classified as turbulence is reassuring and indicates some degree of robustness of the neural network classifier.

\subsection{Evaluation on out-of-distribution data}
\label{s.othertest}

A common problem in Machine Learning is that trained neural networks have good performance when tested on data coming from the same distribution as the training data, but the performance deteriorates when tested on other data (e.g. letters in a completely different font). This may be an especially important factor in the case of physics simulation data, where the parameters of the simulation implicitly define a particular probability distribution, and there is a danger that the neural network will learn the particularities of that simulation instead of learning more universal features.

\begin{table}[ht]
\begin{tabular}{|l|l|}
\hline
Test dataset & accuracy \\
\hline
incompressible fluid, $k_{forcing}=15$, $\nu=10^{-3}$  \textit{(training distribution)} & 100\% \\
\hline
incompressible fluid, $k_{forcing}=7$, $\nu=10^{-3}$ & 100\% \\
\hline
incompressible fluid, $k_{forcing}=12$, $\nu=10^{-3}$ & 100\% \\
\hline
incompressible fluid, $k_{forcing}=26$, $\nu=10^{-3}$ & 100\% \\
\hline
incompressible fluid, $k_{forcing}=15$, $\nu=10^{-4}$ & 100\% \\
\hline
weakly compressible fluid, $k_{forcing}=15$ & 100\% \\
\hline
\end{tabular}
\caption{Test accuracies of a single neural network trained to distinguish chaos from turbulence for an incompressible fluid with $k_{forcing}=15$ and $\nu=10^{-3}$. The test accuracies are evaluated on images of chaos and turbulence coming from simulations with different parameters.}
\label{Tab:accuracy}
\end{table}

In section~\ref{s.adversarial}, we already evaluated a trained neural network on various kinds of data, but there our emphasis was different. Here we would like to summarize these results from the point of view of test accuracy on chaos vs. turbulence data coming from simulations with different parameters or even different dynamics. The results are given in Table~\ref{Tab:accuracy}.

These findings show that despite training the classifier on quite restricted training data, namely vorticity profiles for an incompressible fluid with a specific forcing scale, the trained network correctly distinguishes chaos from turbulence not only for data obtained from the same simulation, but also for data generated with a different forcing scale or value of viscosity, as well as with different dynamic equations (weakly compressible fluid).

\section{Discussion and Outlook}

The phenomenon of turbulence can be seen as a quintessential complex system in Physics. Yet, the intuitive notion of complexity is very difficult to make precise. 
We studied the complexity of turbulence from the point of view of deep neural networks analyzing the images of turbulence. 
Concretely, we analyze the computational effort made by the neural network necessary for distinguishing turbulence from chaos and from a variety of synthetic noise samples and some classes of real-world images. The computational effort is measured as an effective dimension of the neural network's internal feature representations at various depths \cite{complexity}, which can be understood as a  counting the number of independent features that the neural network has to retain in order to distinguish various classes of images e.g. turbulence from chaos. The variation of this number with depth represents the passage from UV to IR (see Section~\ref{s.lincomplexity}) and from elementary to more nonlocal and ``composite'' features. Clearly, the employed methods can be used also in many other contexts.

The key findings of the present paper are as follows: i) distinguishing turbulence from other images is much simpler than distinguishing between real-world images like cats vs. dogs, ii) the pattern of complexity with depth/scale is much more concentrated in the UV in comparison to classifying cats vs. dogs, iii) in the case of turbulence vs. chaos, computation more in the IR (i.e. deeper in the network) is needed, but still much less than for real world images,
iv) we identified evidence that for distinguishing turbulence from chaos, the network uses only the quadratic correlation structure, 
v) when using \textit{untrained} randomly initialized neural networks, the effective dimensions for turbulence are much higher than the ones after training, 
as discussed below.

The above way of looking at deep neural network processing is interesting also from another perspective:
One can view the action of a deep neural network as transforming the original pixel representation into increasingly nonlocal (and nonlinear) representations. This picture has numerous striking analogies in Physics, like the notion of holography (for a review see \cite{Aharony:1999ti}), which can be understood as transforming the original degrees of freedom of a Quantum Field Theory into more and more nonlocal ones and encoding them in a higher dimensional geometry\footnote{See \cite{Hashimoto_2018,Hashimoto_2019} for some concrete analogies.} or the related notion of Renormalization Group.
Consequently, a trained deep neural network can be viewed as a ``holography for probability distributions'' and used to explore the properties of datasets at different length scales rather than just classify them~\cite{complexity}.
From this perspective, the variation of the effective dimensions with depth can be seen as quantifying the ``shape'' (or ``geometry'')  of the feature representations as we pass from the UV to the IR. Hence the findings i)-iii) indicate a marked difference in this respect between distinguishing real-world images and distinguishing even apparently complex Physics-based images. In particular, it suggests a varying structure of correlations between the different parts of images sampled from different datasets. 

Given the overall low effective dimensions for distinguishing turbulence from chaos, one could expect that the deep neural network would perform the classification based just on some relatively uncomplicated features (our finding iv) above). In order to investigate this issue, we constructed proxy images, which shared with turbulence only the quadratic correlation structure at various scales, and ignored all nonlinear (i.e. higher order) correlations in turbulence. These images were still classified very confidently by the neural network as turbulence, thus supporting the hypothesis that the network just used elements of the Fourier spectrum for identifying turbulence.

This result may be considered as providing a word of caution for applications of deep neural networks for identifying some physical states (like turbulence/phase transitions/criticality etc.). It is quite probable that the neural network would make this classification based only on a partial characterization of the given physical state which was simple enough to work for the training data. Hence the definition of the physical state inferred by the neural network from training data may easily be incomplete (e.g. ignoring higher order statistics in turbulence) or may be based on non-universal features.

The present paper concentrated on decision problems i.e. on assessing the ``computational complexity'' for distinguishing turbulence from other images. This gave a method for comparing the difficulties of such classification tasks and examining how their complexity is distributed as a function of scale.
However, the low values of effective dimensions for the turbulence classifiers
do not imply that turbulence images are simple as our finding v) indicates.
The internal representations of the turbulence images in a \emph{random, untrained} neural network have significantly higher intrinsic dimensionalities. This shows that the dataset by itself is not too simplistic, but for the task of distinguishing turbulence from e.g. chaos, the majority of fine details are irrelevant and can be safely dropped leading to much lower dimensions for the \emph{trained} networks.

Finally, we emphasize that throughout the present paper we studied randomly chosen snapshots of either turbulent or chaotic flow ignoring any properties dealing with dynamics in time. Quantifying and defining complexity for the temporal evolution remains thus as a very interesting open problem for future research.

\bigskip

\noindent{}{\bf Acknowledgments.} 
We would like to thank J.R. Westernacher-Schneider for discussions about numerically evolving weakly compressible flows. RJ was supported by the research project \textit{Bio-inspired artificial neural networks} (grant no. POIR.04.04.00-00-14DE/18-00) within the Team-Net program of the Foundation for Polish Science co-financed by the European Union under the European Regional Development Fund and by a grant from the Priority Research Area DigiWorld under the Strategic Programme Excellence Initiative at Jagiellonian University. The work of Y.O. is supported in part by  Israel Science Foundation Center
of Excellence.

\noindent{}{\bf Data availability statement}
The data that support the findings of this study are available from the corresponding author, T.W., upon reasonable request.

\noindent{}{\bf Author contribution statement}
T.W. and R.J. carried out the numerical experiments and analyzed the results. All authors
worked on developing the main idea, discussed the results and contributed to the final manuscript.

\noindent{}{\bf Conflict of interest}
The authors declare that they have
no conflict of interest.

\appendix

\section{Velocity Images}
\label{app:vel}

In the previous sections we calculated the effective using images of the vorticity pseudoscalar.
It is natural to ask whether the same conclusions can be drawn if we use the velocity vector field.
We present results for the effective dimension and adversarial example using images of the $v_x$ and $v_y$
components of the velocities instead of the vorticity. In the incompressible case, the vorticity contains all the information of the velocity field but in the compressible case, the vorticty will miss the irrotational information.
Alternatively, there is a statistical isotropy in the incompressible case, which is broken in the compressible
case.
We generate new Fourier noise using the statistics of the velocity images. 

\subsection{Effective Dimensions}

\begin{figure}[ht]
    \centering
    \includegraphics[width=0.47\textwidth]{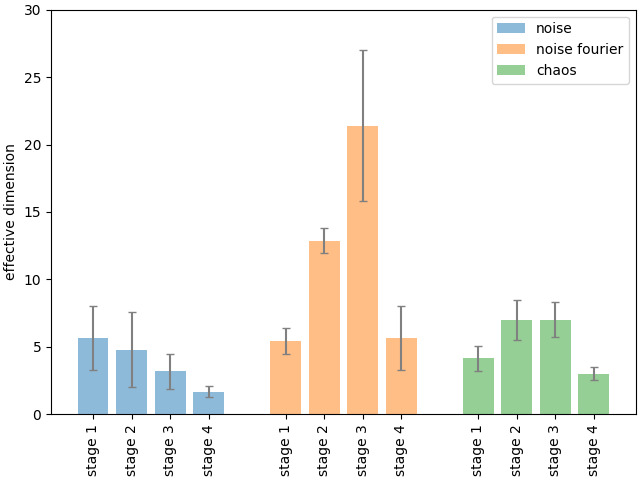}
    \includegraphics[width=0.47\textwidth]{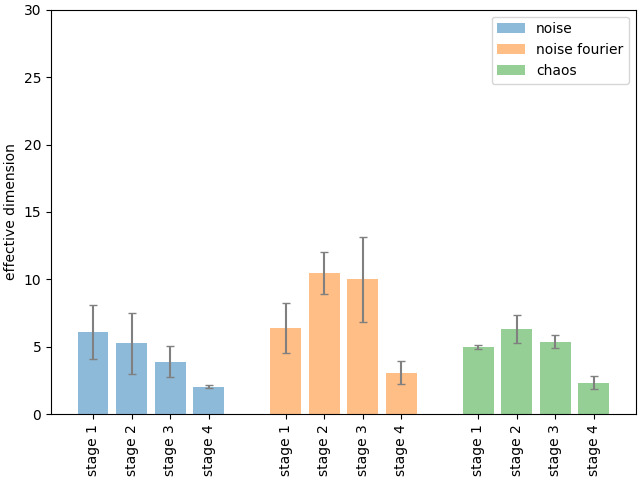}
    \caption{Effective dimensions for classifying weakly compressible turbulent velocities. Left panel is the x component while the right is the y component.}
    \label{fig:vel_summary}
\end{figure}

In Fig.~\ref{fig:vel_summary} , we show the pattern of effective dimensions for classifying turbulence velocity images from chaos as well as the two types of noise discussed in section~\ref{s.noise}. We show the results for the incompressible and weakly compressible case. The effective dimensions are similar to the vorticity ones except the weakly compressible chaos having overall larger effective dimension. 
We see that while in the incompressible case the effective dimensions using $v_x$ or $v_y$ are similar, this is not the case in the compressible case. This can be attributed to the broken statistical isotropy of
the compressible fluid motion.

\bigskip

\begin{figure}[ht]
    \centering
    \includegraphics[width=0.47\textwidth]{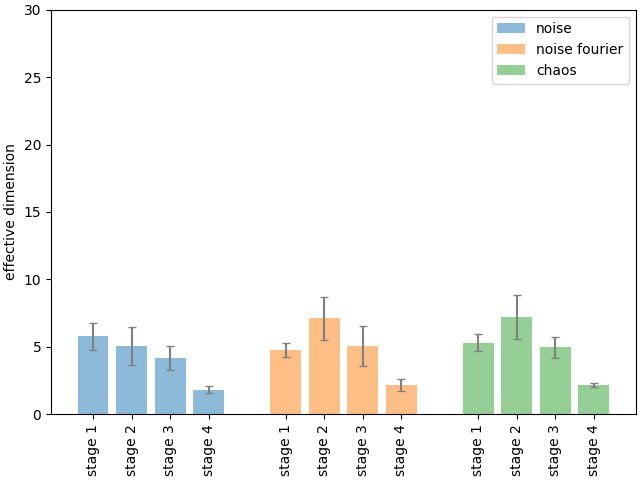}
    \includegraphics[width=0.47\textwidth]{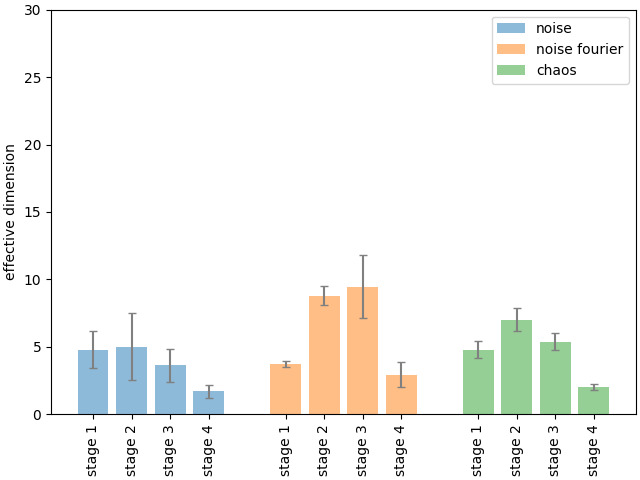}
    \caption{Effective dimensions for incompressible turbulent velocities. Left panel is the x component while the right is the y component.}
    \label{fig:vel_summaryin}
\end{figure}

\bigskip

\subsection{Adversarial examples}

As for the adversarial examples using the velocity data, we recover the same results as the vorticity data. In Table~\ref{Tab:comp_attack_vel} we show the results of the adversarial examples for the weakly compressible case. Table~\ref{Tab:incomp_attack_vel} has the results of the adversarial examples for the incompressible case.

\begin{table}[ht]
\begin{tabular}{l|l|l|}
\cline{2-3}
                                    & Chaos & Turbulence \\ \hline
\multicolumn{1}{|l|}{Weakly compressible Noise Fourier} &   0\%    &      100\%      \\ \hline
\multicolumn{1}{|l|}{Weakly compressible Chaos Noise Fourier} &    100 \%  &       0 \%    \\ \hline
\multicolumn{1}{|l|}{Incompressible Noise Fourier} &  0 \%    &      100\%      \\ \hline
\multicolumn{1}{|l|}{Incompressible Chaos Noise Fourier} &     100\%  &        0\%    \\ \hline
\end{tabular}
\caption{The best case of the compressible velocity classifier.}
\label{Tab:comp_attack_vel}
\end{table}

\begin{table}[ht]
\begin{tabular}{l|l|l|}
\cline{2-3}
                                    & Chaos & Turbulence \\ \hline
\multicolumn{1}{|l|}{Incompressible Noise Fourier} &   0\%    &      100\%      \\ \hline
\multicolumn{1}{|l|}{Incompressible Chaos Noise Fourier} &     100\%  &        0\%    \\ \hline
\multicolumn{1}{|l|}{Weakly compressible Noise Fourier} &   100\%    &      100\%      \\ \hline
\multicolumn{1}{|l|}{Weakly compressible Chaos Noise Fourier} &     100\%  &        0\%    \\ \hline
\end{tabular}
\caption{The best case of the incompressible velocity classifier.}
\label{Tab:incomp_attack_vel}
\end{table}

\bibliographystyle{JHEP}
\bibliography{sample}

\end{document}